\documentclass{article}

 \PassOptionsToPackage{numbers, compress}{natbib}
 \usepackage[final]{nips_2018_MLPCD}

\usepackage[utf8]{inputenc} % allow utf-8 input
\usepackage[T1]{fontenc}    % use 8-bit T1 fonts
\usepackage{hyperref}       % hyperlinks
\usepackage{url}            % simple URL typesetting
\usepackage{booktabs}       % professional-quality tables
\usepackage{amsfonts}       % blackboard math symbols
\usepackage{nicefrac}       % compact symbols for 1/2, etc.
\usepackage{microtype}      % microtypography
\usepackage{color}
\usepackage{subfigure}
\usepackage{graphicx}
\usepackage{tabularx}
\usepackage{multirow}
\usepackage{wrapfig}
\usepackage{algorithm}
\usepackage[noend]{algpseudocode}

\usepackage{fancyhdr, epsfig, epsf, amsmath, amssymb, subfigure,color,amsfonts}
\usepackage{threeparttable}
\usepackage{dsfont}
\usepackage{enumerate}
\usepackage{comment}
\usepackage{multirow,multicol,booktabs}
\usepackage{bigstrut}
\usepackage[most]{tcolorbox}
\tcbuselibrary{breakable} 
\usepackage{enumitem}

\def\({\left(}
\def\){\right)}

\def\[{\left[}
\def\]{\right]}

\title{\fontsize{16}{18}\selectfont Communication-Efficient On-Device Machine~Learning: Federated Distillation and Augmentation \\under Non-IID Private Data}

\author{
  Eunjeong Jeong$^*$, Seungeun Oh$^*$, Hyesung Kim,\\
  \textbf{Seong-Lyun Kim}\\
  Yonsei University\\
  \{\texttt{ejjeong,seoh,hskim,slkim}\}\\\texttt{@ramo.yonsei.ac.kr}
  \and
  \textbf{Jihong Park,}\\ 
  \textbf{Mehdi Bennis}\\
  University of Oulu\\
  \{\texttt{jihong.park,mehdi.bennis}\}\\
  \texttt{@oulu.fi}  
}

\begin{document}

\maketitle

\def\thefootnote{*}\footnotetext{Equal contribution}\def\thefootnote{\arabic{footnote}}

\begin{abstract}
On-device machine learning (ML) enables the training process to exploit a massive amount of user-generated private data samples. To enjoy this benefit, inter-device communication overhead should be minimized. With this end, we propose \emph{federated distillation (FD)}, a distributed model training algorithm whose communication payload size is much smaller than a benchmark scheme, federated learning (FL), particularly when the model size is large. Moreover, user-generated data samples are likely to become non-IID across devices, which commonly degrades the performance compared to the case with an IID dataset. To cope with this, we propose \emph{federated augmentation (FAug)}, where each device collectively trains a generative model, and thereby augments its local data towards yielding an IID dataset. Empirical studies demonstrate that FD with FAug yields around 26x less communication overhead while achieving 95-98\% test accuracy compared to FL.

\end{abstract}
\section{Introduction}

Big training dataset kickstarted the modern machine learning (ML) revolution. On-device ML can fuel the next evolution by allowing access to a huge volume of private data samples that are generated and owned by mobile devices \cite{Google_ai1,Mehdi_spectrum}. Preserving data privacy facilitates such access, in a way that a global model is collectively trained not by directly sharing private data but by exchanging the local model parameters of devices, as exemplified by \emph{federated learning (FL)} \cite{Jakub_FL16,Brendan17,KimWCL:18,FL_v2x,FL_Nishio,DiffFL18,ARM18}.

Unfortunately, in FL, performing the training process at each device side entails communication overhead being proportional to model sizes, forbidding the use of large-sized models. Furthermore, a user-generated training dataset is likely to be non-IID across devices. Compared to its IID dataset counterpart, it decreases the prediction accuracy by up to 11\% for MNIST and 51\% for CIFAR-10 under FL \cite{ARM18}. The reduced accuracy can partly be restored by exchanging data samples, which may however induce an excessive amount of communication overhead and privacy leakage.

On this account we seek for a communication-efficient on-device ML approach under non-IID private data. For communication efficiency, we propose \emph{federated distillation (FD)}, a distributed online knowledge distillation method whose communication payload size depends not on the model size but on the output dimension. Prior to operating FD, we rectify the non-IID training dataset via \emph{federated augmentation (FAug)}, a data augmentation scheme using a generative adversarial network (GAN) that is collectively trained under the trade-off between privacy leakage and communication overhead. The trained GAN empowers each device to locally reproduce the data samples of all devices, so as to make the training dataset become IID.

\section{Federated distillation}
Traditional distributed training algorithms exchange local model parameters every epoch. It gives rise to significant communication overhead in on-device ML where mobile devices are wirelessly interconnected. FL  reduces the communication cost by exchanging model parameters at intervals \cite{Jakub_FL16,Brendan17,KimWCL:18,FL_v2x,FL_Nishio,DiffFL18,ARM18}. On top of such periodic communication, the proposed FD exchanges not the model parameters but the model output, allowing on-device ML to adopt large-sized local models.

The basic operation procedure of FD follows an online version of knowledge distillation \cite{KD14}, also known as \emph{co-distillation (CD)} \cite{OnlineKD}. In CD, each device treats itself as a student, and sees the mean model output of all the other devices as its teacher's output. Each model output is a set of logit values normalized via a softmax function, hereafter denoted as a \emph{logit vector} whose size is given by the number of labels. The teacher-student output difference is periodically measured using cross entropy that becomes the student's loss regularizer, referred to as a \emph{distillation regularizer}, thereby obtaining the knowledge of the other devices during the distributed training process.

CD is however far from being communication-efficient. The reason is because each logit vector is associated with its input training data sample. Therefore, to operate knowledge distillation, both teacher and student outputs should be evaluated using an identical training data sample. This does not allow periodic model output exchanges. Instead, it requires exchanging either model outputs as many as the training dataset size, or model parameters so that the reproduced teacher model can locally generate outputs synchronously with the student model~\cite{OnlineKD}. 

To rectify this, each device in FD stores per-label mean logit vectors, and periodically uploads these \emph{local-average logit vectors} to a server. For each label, the uploaded local-average logit vectors from all devices are averaged, resulting in a \emph{global-average logit vector} per label. The global-average logit vectors of all labels are downloaded to each device. Then, when each device calculates the distillation regularizer, its teacher's output is selected as the global-average logit vector associated with the same label as the current training sample's label.

\begin{figure*}
\subfigure[FD with 2 devices and 2 labels.\label{Fig_FD}]{\includegraphics[width=.5\textwidth]{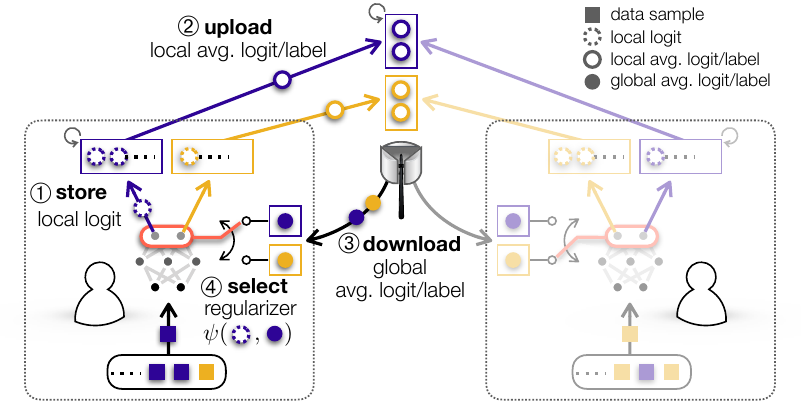}}
\subfigure[FAug with 3 target and 3 redundant MNIST labels.\label{Fig_FAug}]{\includegraphics[width=.5\textwidth]{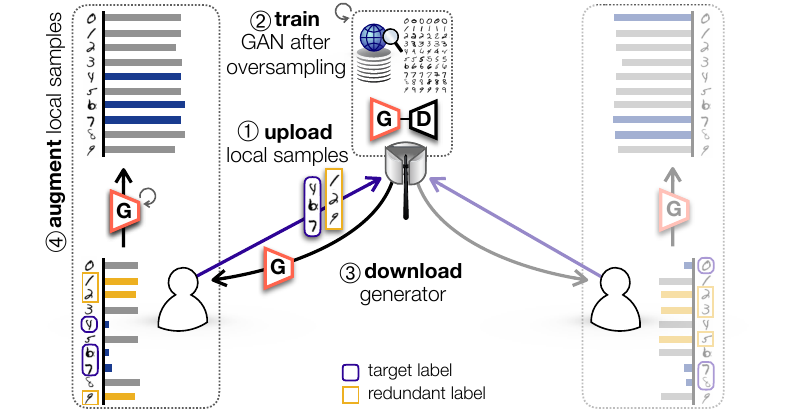}}\vskip -5pt
\caption{\small{Schematic overview of federated distillation (FD) and federated augmentation (FAug).}}
\end{figure*}

The said operation of FD is visualized in Fig.~\ref{Fig_FD}, and is detailed by Algorithm~\ref{euclid}. Notations are summarized as follows. The set $\mathbb{S}$ denotes the entire training dataset of all devices, and $B$ represents the batch of each device. The function $F(w, a)$ is the logit vector normalized by the softmax function, where $w$ and $a$ are the model's weight and input. The function $\phi(p,q)$ is the cross entropy between $p$ and $q$, which is used for both loss function and distillation regularizer. The term $\eta$ is a constant learning rate, and $\gamma$ is a weight parameter for the distillation regularizer. At the $i$-th device, $\bar{F}_{k,\ell}^{(i)}$ is the local-average logit vector at the $k$-th iteration when the training sample belongs to the $\ell$-th ground-truth label, and $\hat{F}_{k,\ell}^{(i)}$ is the global-average logit vector that equals $\hat{F}_{k,\ell}^{(i)}=\sum_{j\neq i}\bar{F}^{(j)}_{k,\ell}/(M-1)$ with a number $M$ of devices, and $cnt_{k,\ell}^{(i)}$ is the number of samples whose ground-truth label is $\ell$.

\begin{algorithm}{\small
	\caption{Federated distillation (FD)}\label{euclid}
	\begin{algorithmic}[1]
	 \Require Prediction function: $F(w, input)$, Loss function: $\phi (F, label)$, Ground-truth label: $y_{input}$
	
  	        \While  {not converged}
	        		\Procedure{Local Training Phase }{at each device}
	        \For  {$n$ steps}
	        : $B$, $y_B$ $\gets$ ${\mathbb{S}}$
	        \For  {sample $b\in B$}
	         \State$w^{(i)} \gets w^{(i)}-\eta \nabla \{ \phi(F(w^{(i)}, b), y_b)+\gamma \cdot \phi(F(w^{(i)}, b), \hat{F}_{k,y_b}^{(i)}) \}$
	        %\State $w_{k+1}^{(i)} \gets w_{k}^{i}-\eta \Delta \{ \phi(F(w_{k}^{i}, b), y_b)+\gamma \cdot \phi(F(w_{k}^{i}, b), y_b) \}$
	        \State  $F^{(i)}_{k,y_{b}} \gets F^{(i)}_{k,y_{b}}+F(w^{(i)},b)$, $cnt_{k,y_{b}}^{(i)} \gets cnt_{k,y_{b}}^{(i)}+1$
	
	        \EndFor 
	        \EndFor

	        \For {label $\ell =1,2,\cdots,L$}	        
	        \State $\bar{F}_{k,\ell}^{(i)} \gets F_{k,\ell}^{(i)} / cnt_{k,\ell}^{(i)}:$ 
	        \Return {$\bar{F}_{k,\ell}^{(i)}$} to server
	        \EndFor   
	        \EndProcedure
		\Procedure  {Global Ensembling Phase } {at the server}
		\For  {each device $i=1,2,\cdots,M$}
		\For  {label $\ell=1,2,\cdots,L$}
		\State  $\bar{F}_{k,\ell} \gets \bar{F}_{k,\ell} + \bar{F}_{k,\ell}^{(i)}$
		\EndFor
		\EndFor
		\For  {each device $i=1,2,\cdots,M$}
	    \For  {label $\ell=1,2,\cdots,L$}
		\State  $\hat{F}_{k+1,\ell}^{(i)} \gets \bar{F}_{k,\ell} - \bar{F}_{k,\ell}^{(i)}$, $\hat{F}_{k+1,\ell}^{(i)} \gets \hat{F}_{k+1,\ell}^{(i)}/(M-1):$ \Return {$\hat{F}_{k+1,\ell}^{(i)}$} to device $i$  
		\EndFor
		\EndFor
		\EndProcedure

		\EndWhile{end while}
		\end{algorithmic}}
	\end{algorithm}

\section{Federated augmentation}
The non-IID training dataset of on-device ML can be corrected by obtaining the missing local data samples at each device from the other devices \cite{ARM18}. This may however induce significant communication overhead, especially with a large number of devices. Instead, we propose FAug where each device can locally generate the missing data samples using a generative model.

The generative model is trained at a server with high computing power and a fast connection to the Internet. Each device in FAug recognizes the labels being lacking in data samples, referred to as \emph{target labels}, and uploads few seed data samples of these target labels to the server over wireless links. The server oversamples the uploaded seed data samples, e.g., via Google's image search for visual data, so as to train a conditional GAN \cite{CondGAN14}. Finally, downloading the trained GAN's generator empowers each device to replenish the target labels until reaching an IID training dataset, which significantly reduces the communication overhead compared to direct data sample exchanges. This procedure is illustrated in Fig.~\ref{Fig_FAug}

The operation of FAug needs to guarantee the privacy of the user-generated data. In fact, each device's data generation bias, i.e., target labels, may easily reveal its privacy sensitive information, e.g., patients' medical checkup items revealing the diagnosis result. To keep these target labels private from the server, the device additionally uploads redundant data samples from the labels other than the target labels. The privacy leakage from each device to the server, denoted as \emph{device-server privacy leakage (PL)}, is thereby reduced at the cost of extra uplink communication overhead. At the $i$-th device, its device-server PL is measured as $|\mathbb{L}^{(i)}_t|/(|\mathbb{L}^{(i)}_t| + |\mathbb{L}^{(i)}_r|)$, where $|\mathbb{L}^{(i)}_t|$ and $|\mathbb{L}^{(i)}_r|$ denote the numbers of target and redundant labels, respectively.

The target label information of a device can also be leaked to the other devices since they share a collectively trained generator. Indeed, a device can infer the others' target labels by identifying the generable labels of its downloaded generator. This privacy leakage is quantified by \emph{inter-device PL}. Provided that the GAN is always perfectly trained for all target and redundant labels, the inter-device~PL of the $i$-th device is defined as $|\mathbb{L}^{(i)}_t|/ \big|\bigcup_{j=1}^M  (\mathbb{L}^{(j)}_t \cup \mathbb{L}^{(j)}_r)\big|$. Note that the inter-device~PL is minimized when its denominator equals to the maximum value, i.e., the number of the entire labels. This minimum leakage can be achieved so long as the number of devices is sufficiently large, regardless of the sizes of the target and redundant labels.

\section{Evaluation}

In this section, we evaluate the proposed FD and FAug under a non-IID MNIST training dataset that is constructed by the following procedure. In the MNIST training dataset with 55,000 samples, we uniformly randomly select 2,000 samples, and allocate them to each device. A set of these 2,000 samples is divided into 10 subsets according to the ground-truth labels. Then, at each device, we uniformly randomly select target labels with a pre-defined number of target labels, and eliminate around 97.5\% of the samples in the target labels such that each target label contains 5 samples.

With this non-IID MNIST training dataset, each device has a 5-layer convolutional neural network (CNN) that consists of: 2 convolutional layers, 1 max-pooling layer, and 2 fully-connected layers. The device conducts its local training with the batch size set as~64. As a benchmark scheme against FD, we consider FL \cite{Brendan17} with or without FAug. In both FD and FL, each device performs 250 local iterations of model updates ($n=250$) before exchanging information with the other devices, constituting one global iteration. The process repeats for a maximum of 16 global iterations.
%In both FD and FL, each device exchanges the desired information at every 250 local iterations ($n=250$), denoted as a global iteration, up to 16 global iterations.
For each global iteration, FD exchanges 100 logits for uplink (i.e., $\{\bar{F}_{k,\ell}^{(i)}\}_{\ell=1}^L$)
and another 100 logits for downlink (i.e., $\{\hat{F}_{k,\ell}^{(i)}\}_{\ell=1}^L$), each of which comprises 10~logit vectors containing 10~elements. On the other hand, FL exchanges the CNN's 1,199,648 model parameters per global iteration equally for both uplink and~downlink.

In FAug, the server has a conditional GAN consisting of a 4-layer generator neural network and a 4-layer discriminator neural network. When downloading a trained generator, the communication overhead is proportional to the generator's model size, given as 1,493,520 parameters.

\begin{table*}[t]{
\centering
\resizebox{\textwidth}{!}{\begin{tabular}{c c c c c c c c c c} 
    \toprule
   \multirow{2}[4]{*}{ Methods} & \multicolumn{5}{c}{Accuracy w.r.t. the number of  devices}&\multicolumn{4}{c}{Communication cost} \\ 
    \cmidrule(rl){2-6}  \cmidrule(rl){7-10}
    & 2  & 4 & 6 & 8 & 10 & Logits &Model parameters & Samples & Total (bits) \\ 
    \cmidrule(r){1-1}\cmidrule(l){2-6}   \cmidrule(l){7-9}\cmidrule(l){10-10}
    \multicolumn{1}{l}{FD + FAug}&0.8464& 0.8526& 0.8498&  0.8480& \textbf{0.8642}  &3,200&1,493,520 & 15 & \textbf{47,989,120} \\
    \multicolumn{1}{l}{FD (non-IID)}&0.7230& 0.7304 & 0.6951&  0.6839& \textbf{0.7524}&3,200&-&-& \textbf{102,400} \\ \cmidrule(l){1-10} 
    \multicolumn{1}{l}{FL + FAug} & 0.9111& 0.8654& 0.8956&  0.9101& \textbf{0.9259} &-&39,882,256&15&\textbf{1,276,326,272} \\
    \multicolumn{1}{l}{FL (non-IID)} &0.8077& 0.8684 & 0.8738&  0.8878& \textbf{0.9060} &-&38,388,736&-&\textbf{1,228,439,552}  \\
    \bottomrule
\end{tabular}}
\caption{ Test accuracy and communication cost (3 target labels, no redundant label).}\label{tab:1}\vspace{-10pt}}
\end{table*}

\begin{figure}[t]
   \centering\
   \subfigure[{Accuracy per label.}\label{acc_per_label}]{\includegraphics[width=0.325\textwidth]{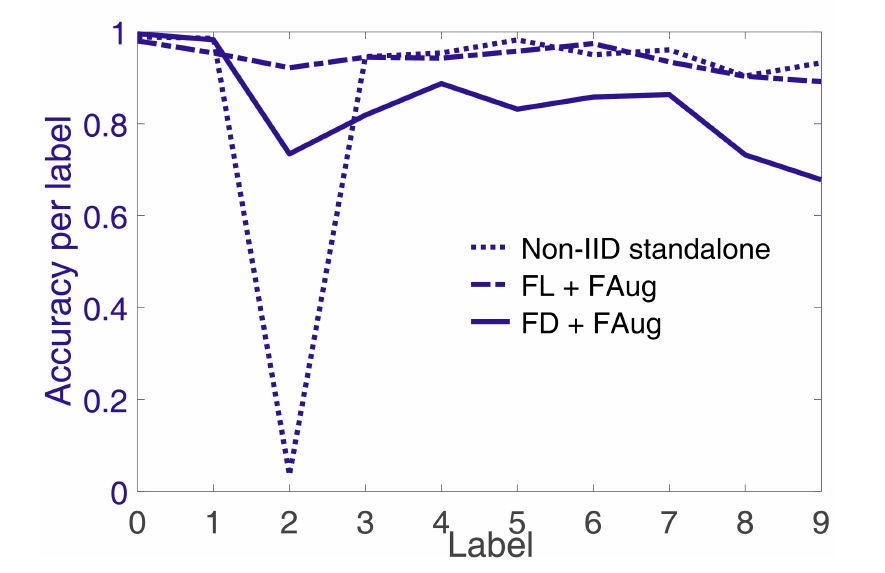}}
   \subfigure[{Inter-device PL and accuracy.}\label{acc_ueue_p}]{\includegraphics[width=0.325\textwidth]{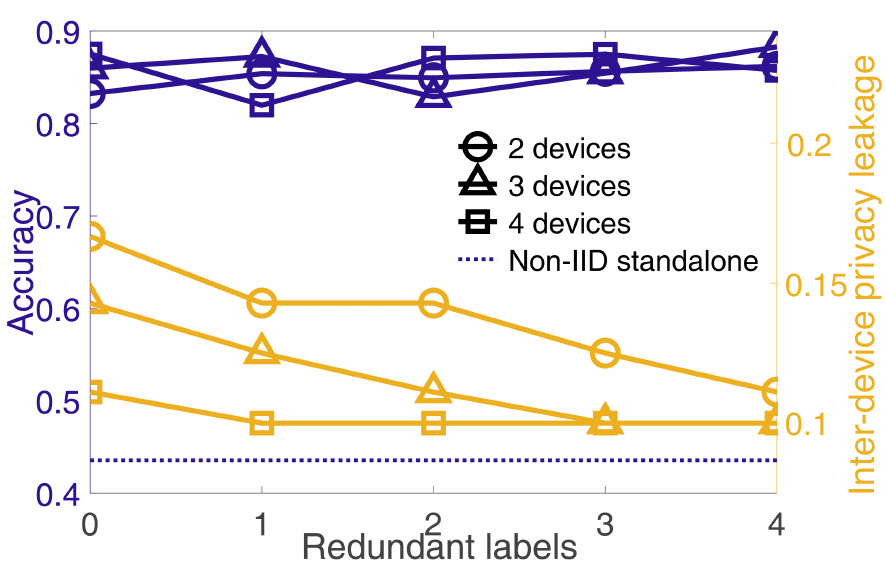}}   
      \subfigure[{Device-server PL.}\label{acc_bsue_p}]{\includegraphics[width=0.325\textwidth]{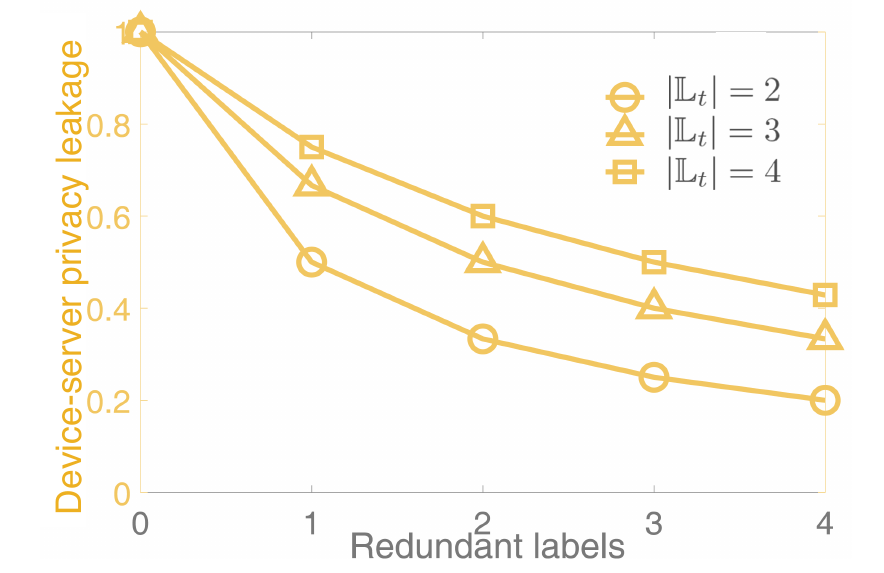}}   
   \caption{Test accuracy and privacy leakage (PL) under a non-IID MNIST dataset: (a) accuracy per label under FL or FD with FAug, compared to the non-IID standalone case; (b) inter-device PL and accuracy under FD with FAug; and (c) device-server PL under FAug for different numbers of target~labels.}
\end{figure}

At a uniformly randomly chosen reference device, Table \ref{tab:1} provides the test accuracy and communication cost of FD and FL, with or without FAug. For FD, the communication cost is defined as the number of exchanged logits, whereas the cost for FL is given as the number of exchanged model parameters. With FAug, the communication cost comprises the number of uploading data samples and the number of downloading model parameters of the trained generator. For each method, the total communication cost is evaluated by considering that each pixel of MNIST sample (28x28 pixels) occupies 8 bits, while each logit and each model parameter equally consume 32 bits.

In Table \ref{tab:1}, compared to FL, we observe that FD significantly reduces the communication cost while slightly compromising the test accuracy. The reduced accuracy of FD can be complemented by FAug, without incurring severe communication overhead. To be specific, compared to FL with a different number of devices, FD achieves 77-90\% test accuracy that can be improved by FAug as 95-98\% test accuracy. The aggregate communication cost of FD and FAug is still around 26x smaller than the cost of FL. Note that in return for the communication overhead, FL is more robust against non-IID dataset. This is observed via FAug that increases the test accuracy of FD by 7-22\% and the accuracy of FL by 0.8-2.7\%, compared to the cases without FAug, i.e., non-IID~dataset.

Fig. \ref{acc_per_label} illustrates the per-label test accuracy when `2' is the target label. For the target label, the standalone training of the reference device under the original non-IID dataset yields 3.585\% test accuracy, which is increased via FAug with FD or FL by 73.44\% or 92.19\%, validating the effectiveness of FAug in combination with both FD and FL. For the entire labels, Fig. \ref{acc_ueue_p} shows that FD with FAug achieves around 2x higher test accuracy compared to the standalone training, regardless of the number of devices and of the number of redundant labels. For more devices or redundant labels, the inter-device PL decreases, since both aspects respectively increase the denominator $\big|\bigcup_{j=1}^M  (\mathbb{L}^{(j)}_t\cup \mathbb{L}^{(j)}_r)\big|$ of the inter-device~PL. Similarly, Fig. \ref{acc_bsue_p} describes that the device-server PL decreases with the number of redundant labels, and increases with the number of target labels.

\section{Concluding remarks}
Towards enabling on-device ML, we introduced FD and FAug that are communication-efficient training and data augmentation algorithms, respectively. Empirical studies showed their effectiveness that achieves comparably high accuracy with much smaller communication overhead compared to~FL.

In future work, the performance of FD can further be improved with a slight modification. In fact, model output accuracy increases as the training progresses. During the local logit averaging process, it is thus better to take a weighted average in a way that the weight increases with the local computation time. Furthermore, FD and FL can be combined towards balancing communication-efficiency and accuracy. As an example, one can exploit FD in the uplink and FL in the downlink, on the ground that the downlink wireless communication links are commonly faster than the uplink~\cite{JHParkTWC:15}. Lastly, using the differential privacy framework \cite{DiffFL18}, the privacy guarantee of FAug can be ameliorated by inserting a proper amount of noise into the uploading seed data samples, which is an interesting topic for future research.

\section*{Acknowledgement}
This research was supported in part by a grant to Bio-Mimetic Robot Research Center Fonded by Defense Acquisition Program Administration, and by Agency for Defense Development (UD160027ID), in part by the Academy of Finland project CARMA, and 6Genesis Flagship (grant no. 318927), in part by the INFOTECH project NOOR, in part by the Kvantum Institute strategic project SAFARI, and in part by Basic Science Research Program through the National Research Foundation of Korea (NRF) funded by the Ministry of Science and ICT (NRF-2017R1A2A2A05069810).

\nocite{*}
\bibliographystyle{unsrt}

\end{document}